\documentclass[a4paper, 10pt, conference]{ieeeconf}      

\IEEEoverridecommandlockouts                              

\title{\LARGE \bf
When Would You Trust a Robot? A Study on Trust and Theory of Mind in Human-Robot Interactions}

\author{Wenxuan Mou$^{1}$*, Martina Ruocco$^{1}$*, Debora Zanatto$^{2}$ and Angelo Cangelosi$^{1}$ \\
$^{1}$Department of Computer Science, University of Manchester, UK\\
$^{2}$Department of Computer Science, University of Bristol, UK
 \thanks{* {\tt\small for equal contributions in an alphabetical order. }}
}

\usepackage{flushend}
\usepackage{graphicx}
\usepackage{booktabs}
\usepackage{multirow}
\usepackage{xcolor}
\usepackage[switch]{lineno}
\usepackage{subcaption}
\usepackage{makecell}

\definecolor{amaranthred}{rgb}{0.83,0.13,0.18}

\begin{document}


\maketitle
\thispagestyle{empty}
\pagestyle{empty}

\begin{abstract}
Trust is a critical issue in human–robot interactions (HRI) as
it is the core of human desire to accept and use a non-human agent.
Theory of Mind (ToM) has been defined as the ability to understand the beliefs and intentions of others that may differ from one’s own.
Evidences in psychology and HRI suggest that trust and ToM are interconnected and interdependent concepts, as the decision to trust another agent must depend on our own representation of this entity's actions, beliefs and intentions.
However, very few works take ToM of the robot into consideration while studying trust in HRI. 
In this paper, we investigated whether the exposure to the ToM abilities of a robot could affect humans' trust towards the robot. To this end, participants played a Price Game with a humanoid robot (Pepper) that was presented having either \textit{low-level ToM} or \textit{high-level ToM}. Specifically, the participants were asked to accept the price evaluations on common objects presented by the robot. The willingness of the participants to change their own price judgement of the objects (i.e., accept the price the robot suggested) was used as the main measurement of the trust towards the robot. 
Our experimental results showed that robots possessing a high-level of ToM abilities were trusted more than the robots presented with low-level ToM skills.
\end{abstract}


\section{INTRODUCTION} \label{se:intro}

The field of Human-Robot Interaction (HRI)  aims to investigate and improve the communication and performance between robots and humans~\cite{vinanzi2019would, zanatto2019investigating, romeo2019deploying}. 
Indeed, with the increasing presence of robots in several domains and their expanding availability to non-expert users (e.g., industrial applications, healthcare assistance, and education environments), establishing mutual trust has become a critical issue in HRI. As several authors already stated, trust is a necessary component for building and maintaining a proper social relationship between humans and robots~\cite{hancock2011can, breazeal2005effects}. In particular, it is one of the basic prerequisites for the construction of a cooperative environment, which, in turn, is the final aim of HRI. Only when humans trust the robots, in fact, it becomes possible to establish the cooperation between the two parties. 

Investigations of the factors affecting trust in HRI have listed a large number of determinants, from physical features to specific socio-cognitive abilities \cite{hancock2011meta}. Over the last decades, with the increasing usage of robots in social contexts, research has investigated the role of the robot social skills on the human perception of the robots and how those competencies could affect or even predict how humans would interact with them \cite{salem2015would, shu2018human}. Furthermore, the robot social skills have also been used as a key for understanding the human expectations. In particular, recent interest have been dedicated on what specific social cues could be involved in developing and maintaining trust in HRI~\cite{thesisDebora}. For example, it has been found that people trust more a humanoid robot (iCub) when it engages in a more human-like social gaze behaviour, compared to the use of a fixed gaze~\cite{thesisDebora}. Also, Ham et al.,~\cite{ham2011making} found that the robot gaze increased its persuasive capabilities. Furthermore, Huang and Thomaz reported that robots performing joint attention were also rated as more competent~\cite{huang2010joint}.

Nevertheless, gaze and joint attention are not the only useful scaffold that designer can use to improve trust in HRI. Rather, also showing the ability to emphasise with others and understanding their intentions, beliefs or emotional states could have an impact on trust. This kind of ability goes under the name of Theory of Mind. Specifically, it is the ability to attribute mental states (such as beliefs, intentions and desires) to others that may differ from one's own~\cite{leslie1987pretense}. Such ability enables us humans to connect and empathise with others' states, and is also important for establishing a trustworthy social relationship in the Human-Human Interactions (HHI). Vanderbilt et al.~\cite{vanderbilt2011development} found that children's ability to distinguish whether a person was trustworthy or not was associated with their ToM ability.
As embodied artificial agents, robots are designed to interact or cooperate with other agents, either other robots or humans.
A robot equipped with ToM ability would be able to estimate the goals and desires of the others and react to others' more quickly and more accurately~\cite{scassellati2002theory}. This, in turn, may increase human trust to the robot and establish the cooperation or interactions between human and the robot more properly. 
Vinanzi et al.~\cite{vinanzi2019would} reported that the ToM capability of a robot is closely related to the abilities of the robot to distinguish reliable sources from unreliable ones. This work focused on the study of whether the ToM capabilities of the robot would affect the ability to find trustworthy person or resources. However, to the best of our knowledge, there is no work exploring how ToM abilities expressed by a robot are related to humans' trust towards the robot.

In this paper, our aim is to investigate whether the capabilities of the robot's Theory of Mind would affect humans' trust towards the robot. To this end, a Price Game developed by Rau et al. \cite{rau2009effects} was used and played between human participants and a humanoid robot (Pepper). In the game, participants had to choose the price of several objects, where Pepper could express its agreement or not. In case Pepper disagreed, participants faced the chance of aligning with the robot judgement or keeping their own choice. Prior the experiment, the participants watched a video that presented the ToM abilities of Pepper robot. It was hypothesised that a robot introduced with higher level of ToM abilities (to be referred as \textit{high-level ToM}) would enhance higher trust over the robot presented with lower level of ToM skills (to be referred as \textit{low-level ToM}). Therefore, participants would be more willing to change their judgements on the Price Game after being introduced to a high-level ToM robot.

The rest of the paper is organised as follows: related works are reviewed in Section~\ref{se:relatedwork}; the proposed methodology is illustrated in Section~\ref{se:method}; the experiments and results are presented and discussed in Section~\ref{se:experiment}; conclusions and future work are described in Section~\ref{se:conclusion}.


\section{RELATED WORK} \label{se:relatedwork}

\subsection{Trust in human-robot interaction} \label{se:rw_trust}

Trust is a fundamental ingredient and an unavoidable dimension that affects human social interactions~\cite{mayer1995integrative}. Moreover, trust is a key factor that affects team dynamics and collaboration, which often involves interdependences among people. Therefore, team members must depend on each other to accomplish their personal and common goals~\cite{mayer1995integrative}. 
Nevertheless, trust is a critical factor and an essential component also in HRI. Humans are expected to work with robots in various scenarios in the future and therefore mutual trust is an essential aspect in increasing the usage of robots and ensuring a successful HRI in different social applications.
Trust in HRI is also influenced by a number of factors. In a meta-analysis, Hancock et al.~\cite{hancock2011meta} identified and analysed three trust-affecting factors in detail, human-related factors, robot-related factors and environmental factors. 

Human-related factors included ability-based features (e.g., attentional capacity and expertise) and characteristics (e.g., personality traits and attitudes toward robots). Little evidence was found for the effect of this source~\cite{hancock2011meta, schaefer2013perception}. This is an important result for our study, since it minimises the influences of the participants' personality traits. The environmental factors resulted in having a moderated effect, and those included team collaboration features (e.g., culture communications) and tasking features (e.g., task type, physical environment). For instance, Lee and See~\cite{lee2004trust} showed that trust was less important when it was in a fixed and well structured environment versus in a more dynamic environment.  
Compared to the environmental factors and human-related factors, Hancock et al.~\cite{hancock2011meta} reported that the robot-related factors played the most important role in the development of trust in HRI. Robot-related factors were divided into performance-based (e.g., reliability and level of automation) and attribute-based factors (e.g., social competences, and anthropomorphism).
These findings are supported by a vast list of studies.
For example, Salem et al.~\cite{salem2015would} showed participants' trust judgements on a  robot were affected by its performance. In particular, participants rated the robot that always performed correctly as more trustworthy than a robot that made mistakes. Similarly, Desai et al.~\cite{desai2012effects} found that people would trust a robot system less when its reliability in autonomous mode decreased. Moreover, it is also reported that the social skills and emotional behaviours expressed by a robot affect humans trust towards the robot~\cite{lohani2016social}. Overall, the recent findings in literature suggest that the robot performance and its social competencies have an important effect on the human trustworthiness perception of the robot.
Importantly, among all the social competencies that a robot should embed to improve trust in HRI, ToM plays a significant role.
Therefore, in this paper we will investigate how this robot-related factor could be an expression of trust in HRI and whether it could influence it.

\subsection{Theory of Mind} \label{se:rw_tom}

Among the ones that tried to design a coherent framework that would represent correctly a human ToM (e.g. \cite{bellas2010multilevel, blum2018simulation, chella2006cognitive, augello2018social, lazzeri2018designing} ) we decided to follow Leslie's model (presented in~\cite{scassellati2002theory}) as it is better suited for human-robot social interaction purposes.
According to Leslie, the world manifests itself to the humans through three classes of events: \textbf{mechanical}, \textbf{actional} and \textbf{attitudinal}.
Furthermore, Leslie reported that a  complete ToM framework is composed of three domain specific modules that deal with each of these classes of events. However, these modules are not innate, rather they develop over time \textit{incrementally} (i.e., one after other) during childhood.

\begin{itemize}
    \item The Theory of Body mechanism deals with the baby's understanding of the physical worlds (\textbf{mechanical} layer, i.e., physical causality between themselves and the surroundings).
    \item The Theory of Mind system-1 mechanism deals with the correlation between agents and the goal-directed actions that they perform (\textbf{actional} layer).
    \item The Theory of Mind system-2 mechanism deals with the representation of other agent's beliefs and mental states (\textbf{attitudinal} layer).
\end{itemize}


One of the earliest psychology tests for the development of Theory of Mind system-2 in children (between 3 and 5 years) is the Sally-Anne test developed by Baron-Cohen et al. \cite{baron1985does}. The classic version is either shown as a cartoon or acted out with dolls. Children are shown a game played by two girls, Sally and Anne. During the game, Sally puts a ball into her basket and then leaves the room. After that, Anne takes the ball from the basket and places it in a box. The child, who watches everything happening, is then asked where Sally would look for the ball when she comes back. The children that have a maturer ToM would forecast that Sally believes the ball is still in the basket, while the children with an immature ToM  would answer from their own perspective, i.e., believe that Sally also knows the ball is already in the box. This is because the children with undeveloped ToM could not understand yet that people have different minds and beliefs from their own.

In this paper, we refer to Leslie's model as our ToM framework, but we simplify the notation, referring only to low-level ToM (comparable to the actional layer) and high-level ToM (comparable to the attitudinal layer). This is due to our experimental aim focusing only on the robot ability to read others' beliefs, which will be based on the Baron-Cohen experimental results. The ToM ability enables humans to understand and empathise with others' states in HHI, which is also important in HRI. Sturgeon et al~\cite{sturgeon2019perception} studied whether a robot demonstrating a ToM influencing human perception of social intelligence in a HRI using questionnaire analysis. Different from their study, in this paper, we focus on investigating how ToM abilities expressed by a robot are related to humans’ trust towards the robot.

\section{METHODS} \label{se:method}
\subsection{Participants}
Thirty-two students and staff members (17 male, 15 female) have been recruited from different departments of the University of Manchester. Recruitment has been mainly performed using email advertisements and flyers. Each participant was randomly assigned to one of the two experimental conditions, i.e., either watch a video showing the robot with high-level ToM or low-level ToM.  

We used G Power 3.1~\cite{faul2009statistical}  to determine that a sample size of twenty-seven participants would
provide 80\% statistical power for detecting a medium-sized
effect equivalent to what we observed in Zanatto~\cite{zanatto2019generalisation} study
($r^2$ = 0.46), assuming a two-tailed t-test and an alpha level
of 0.05. 
All participants were naive as to the purpose of
the investigation and gave informed written consent to participate
in the study.

\subsection{Theory of Mind Videos} \label{se:tom_videos}

Our aim was to investigate whether the capabilities of the robot's Theory of Mind would affect humans' trust towards the robot. To this end, a Price Game developed by Rau et al.~\cite{rau2009effects} was used and played between human participants and a humanoid robot (Pepper). Prior the experiment, the participants watched a video that presented the ToM abilities of the Pepper robot. 

\begin{figure*}[h]
		  \centering
		  \includegraphics[width=1\textwidth]{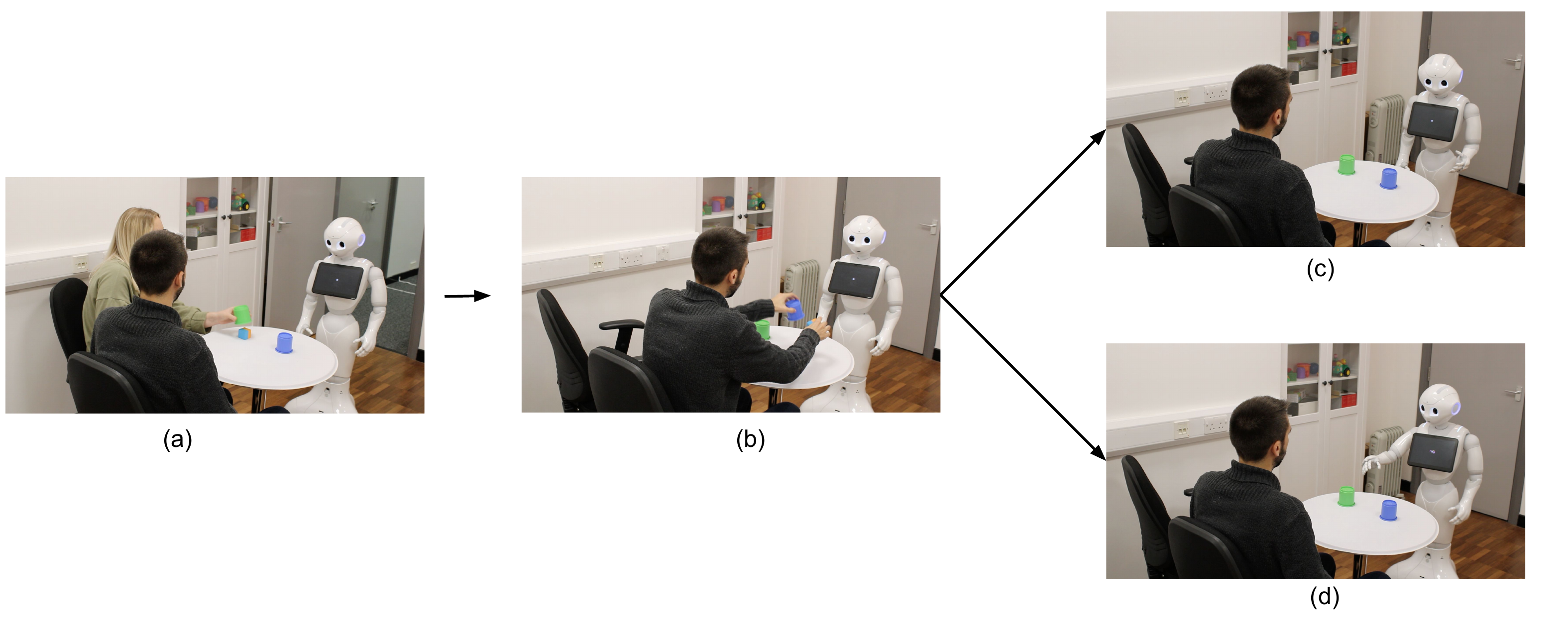}	 \caption{ToM videos. (a) Anne puts the cube under the green cup. (b) Anne leaves the room and Sam puts the cube under the blue cup. (c) The robot is asked where Anne would look for the cube after she comes back. The robot points to the blue cup and says that she would look for the cube under the blue cup, in the \textit{low-level ToM} condition. (d) The robot is also asked where Anne would look for the cube. The robot points to the green cup and say that Anne would look for the cub under the green cup as she did not see Sam moved the cube, in the \textit{high-level ToM} condition.  }
		  \label{fig:ToMvideo}
	    \end{figure*}

The video showed a modified version of the Sally-Anne false-belief task (see section \ref{se:rw_tom} in Related Works). In the video, two people (a male called Sam and a female called Anne) seated in front of the Pepper robot, where a table was positioned between the three parties (Figure \ref{fig:ToMvideo}). On the table were positioned three objects, a cube and two cups (a green and a blue one). Anne put the cube under the green cup and left the room. After that Sam moved the cube under the blue cup. The robot was then asked where Anne would believe the cube was once back in the room. The robot response was different depending on the ToM condition. In the low-level ToM, the robot would respond incorrectly, reporting that Anne would believe the cube was under the blue cup both using gestures (pointing to and look at the blue cup) and speaking. In high-level ToM, the robot would give the correct answer, saying Anne would believe the cube was under the green cup, and at the same time pointing to and looking at the green cup. The detailed scripts for the two videos are presented in the Appendix.

\subsection{Price Game and Procedure}
\begin{figure*}
		  \centering
		  \includegraphics[width=\textwidth]{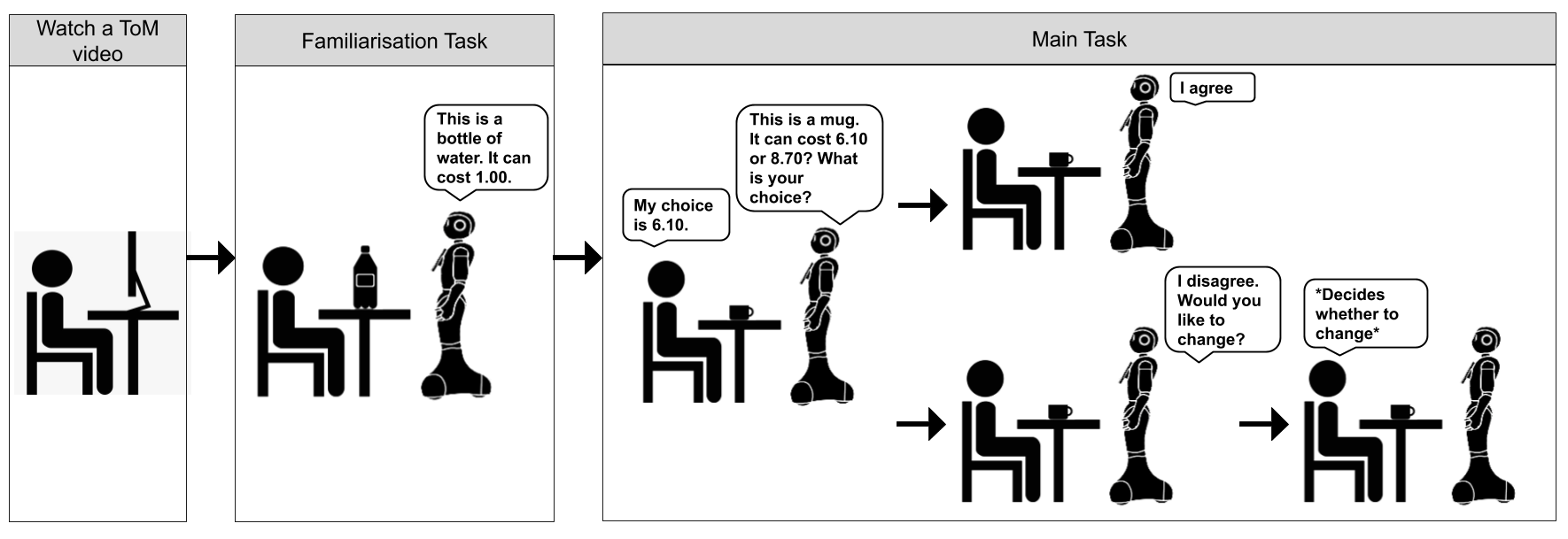}\caption{Illustration of the experimental procedures. Each participant first watched one of the two ToM videos. The participants then started the familiarisation task with the Pepper robot. After that, the participants played the main task of the Price Game, where the robot would provide two prices for each object and the participants were asked to select one of them. The robot could agree or disagree with the participants. If the robot disagreed, the participants had the opportunity to change the price.}
		  \label{fig:flowchart}
	    \end{figure*}

\begin{figure}[h]
		  \centering
		  \includegraphics[width=0.5\textwidth]{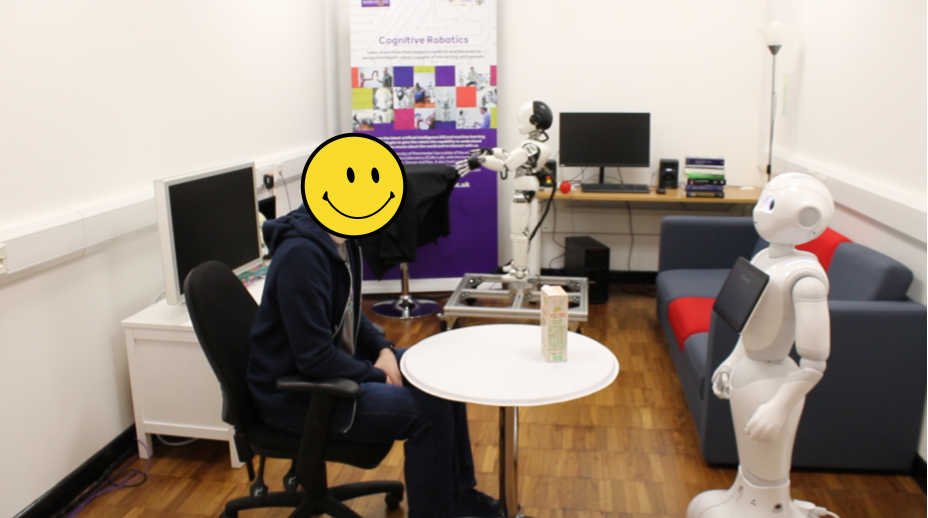}	 \caption{Illustration of the experimental setup. The experiment was conducted in a robotic lab. The participant was always sitting in front of a table facing the robot. }
		  \label{fig:experiment}
	    \end{figure}

Prior to the experimental session, the participants read and signed the information sheet and the consent form. The experiment procedure is shown in Figure \ref{fig:flowchart}. Generally, the experiment included three parts, starting with watching a ToM video, following a familiarisation task and a main task of the Price Game as shown in Figure \ref{fig:experiment}. During the experiment, the participants were seated in front of a table facing the robot. 
	    
\textbf{Watch a ToM video.} The experiment was conducted using a between-subject setup. Each participant only watched one of the videos presenting the ToM abilities of the robot, i.e., either low-level ToM or high-level ToM. After that the participant played the Price Game with Pepper robot. 

\textbf{Familiarisation task.} In the familiarisation session participants did not value the object, whereas only watched the robot pricing the objects. Firstly, an experimenter would place the object on the table; secondly, the robot would look at the object and then moved its gaze to the participant to provide a brief verbal description of the object in terms of its colour, usage, etc; finally, the robot voiced the price of the object. This process got repeated six times in total, right before proceeding with the main task. \\

\textbf{Main task.} For the main task of the Price Game, we followed the procedure used by Zanatto \cite{deborathesis}. Firstly, an experimenter would place the object on the table and the robot looked at the object and moved its gaze to the participant to provide a brief description of the object, which was the same as in the familiarisation task. Secondly, the robot provided two prices for the object and the participant was asked to select one from these two. The two prices were also displayed on the tablet of Pepper to help the participants to memorise the numbers. After the participant made a decision, the robot voiced its agreement or disagreement with the selected price. In the case that the robot agrees, the robot would say ``I agree''. In the case that the robot disagreed with the participant, the robot would say ``I disagree, would you like to change?''. If the robot disagreed, the participant had to decide whether to change the choice.

In total, 22 common objects were used for each participant. Six of those objects were used for a familiarisation session prior the main task. The other 16 objects were used for the main task session, where the robot was programmed to agree with the participants on the price of 8 objects, while it would disagree on the remaining 8 objects. The robot was programmed to always agree/disagree for the same objects (e.g., the robot would always disagree on the price of the gas refill and always agree on the cereal bowl price, regardless of the participant's choice). The order of the objects presented to the participants followed two different scripts. Each participant was randomly assigned to one of the two versions.

At the end of the Price Game, the participants were asked to fill four questionnaires on a laptop computer. These questionnaires were used as a secondary measure to the main experiment task. Specifically, three short scales
measured Likeability~\cite{reysen2005construction}, Trust~\cite{rau2009effects} and Credibility~\cite{mccroskey1981ethos}. In addition, the Godspeed questionnaire~\cite{bartneck2009measurement} was used to measure a range of HRI factors (anthropomorphism, animacy, likeability,
perceived intelligence and perceived safety). The questionnaire was followed by an interview in which the instructor/experimenter invited participants to describe and comment their experience during the experiments. After the interview, participants were debriefed and dismissed. The experiment took approximately 30 minutes.

\section{Experimental results and analysis} \label{se:experiment}

\subsection{Trust rate}
\begin{figure}
		  \centering
		  \includegraphics[width=0.5\textwidth]{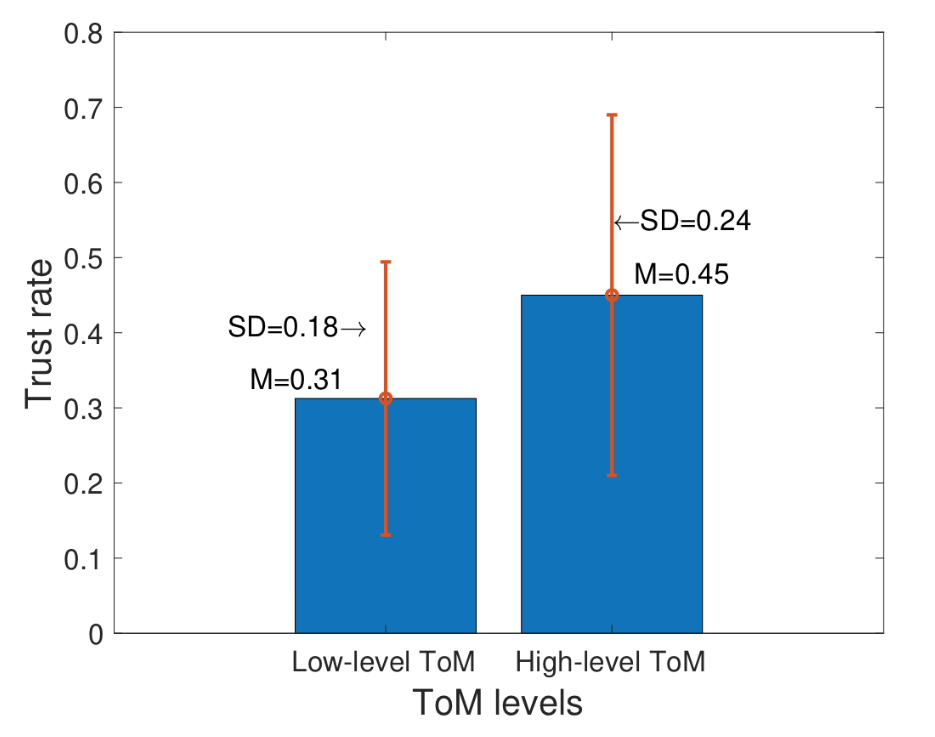}	 \caption{The average trust rate of participants  under two different conditions of ToM capabilities, i.e, low-level ToM and high-level ToM. }
		  \label{fig:trustrate}
	    \end{figure}

We collected data from 32 participants. However, three did not comply with the experimental procedure and were thus excluded from the data analyses.
Participants' willingness to change their choice when the robot disagreed was used as a measurement of trust towards the robot. Specifically, this has been defined as 'trust rate' ($TR$) and has been calculated as follows:
\begin{equation}
    TR = \frac{N_{change}}{N_{robot\_disagree}}
\end{equation}
The average trust rate of all the participants for the two conditions (i.e., high-level ToM and low-level ToM) is shown in Figure \ref{fig:trustrate}.
An independent-samples t-test was performed to compare the participants' trust rate to the robot in high-level ToM and low-level ToM conditions. From the results it emerged that the trust rate was statistically different between the two groups. Participants changed their decision less often in the low-level ToM (Mean = 0.31, SD = 0.18) than in the high-level ToM (Mean = 0.45, SD = 0.24) conditions, $t(27) = 1.73, p = 0.047, d= 0.65$. 
This result suggests that the ToM capabilities presented by the robot in the videos do have an effect on the participants' trust towards the robot. Specifically, our results suggest that when the robot presents higher level of ToM capabilities, humans' trust towards the robot increases.

\subsection{Questionnaire Results}
Questionnaires were used as secondary measure to the main experiment task. For each scale, Z-score normalisation was applied for individual question among all participants and then t-tests were performed to assess the role of the video on the participants' judgement. Results are reported in Table \ref{table:q1} and \ref{table:q2}.
A significant difference between the two types of video has been found for the Credibility scale. Moreover, the Godspeed Questionnaires reported a significant effect of the type of video for Anthropomorphism and Safety scales. For all those scales, participants' ratings were higher in the high-level ToM than the low-level ToM condition.

\begin{table}[]
\caption{Results for Likeability, Trust and Credibility ratings.Two-sample t-tests have been performed to compare participants' rating on the two different ToM videos. For each scale, Mean, SD, t value, and p-value are reported.}
\label{table:q1}
\scalebox{0.95}{
\begin{tabular}{|l|l|l|l|l|l|}
\hline
Dependent                  &  &  &  &                      &                       \\
 Variable                 & ToM Videos & Mean & SD & t                      & $p$                      \\ \hline
\multirow{2}{*}{Likeability}        & low-level ToM     & -0.0195     &1.032    & \multirow{2}{*}{0.871} & \multirow{2}{*}{$> .050$} \\ \cline{2-4}
                                  & high-level ToM     & 0.0182     & 0.941   &                        &                        \\ \hline
\multirow{2}{*}{Trust} & low-level ToM     &  0.005   & 1.117   & \multirow{2}{*}{-0.309} & \multirow{2}{*}{$> .050$} \\ \cline{2-4}
                                  & high-level ToM     & -0.005    & 0.851   &                        &                        \\ \hline
\multirow{2}{*}{Credibility}       & low-level ToM     & -0.076    &  0.998  & \multirow{2}{*}{3.115} & \multirow{2}{*}{$=.003*$} \\ \cline{2-4}
                                  & high-level ToM     & 0.071     & 0.968   &                        &                        \\ \hline
\end{tabular}}

\end{table}


\begin{table}[]
\caption{Results for Godspeed Questionnaire ratings, which includes Anthropomorphism, Animacy, LIkeability, Intelligence and Safety. Two-sample t-tests have been performed to compare participants' rating on the two different ToM videos. For each scale, Mean, SD, t value, and p-value are reported.}
\label{table:q2}
\scalebox{0.95}{
\begin{tabular}{|l|l|l|l|l|l|}
\hline
\makecell{Dependent \\ Variable}                & ToM Videos & Mean & SD & t                      & $p$                      \\ \hline
\multirow{2}{*}{\makecell{Anthropo-\\ morphism}}        & low-level ToM     & -0.105     &0.921    & \multirow{2}{*}{1.937} & \multirow{2}{*}{$=.044*$} \\ \cline{2-4}
                                  & high-level ToM     & 0.098     & 1.040   &                        &                        \\ \hline
\multirow{2}{*}{Animacy} & low-level ToM    &  0.097    & 0.974   & \multirow{2}{*}{-2.096} & \multirow{2}{*}{$>.050$} \\ \cline{2-4}
                                  & high-level ToM     & -0.091     & 0.993   &                        &                        \\ \hline
\multirow{2}{*}{Likeability}       & low-level ToM     & 0.133    &  0.905  & \multirow{2}{*}{-3.932} & \multirow{2}{*}{$> .050$} \\ \cline{2-4}
                                  & high-level ToM     & -0.124     & 1.047  &                        &                        \\ \hline
\multirow{2}{*}{Intelligence}       & low-level ToM     & -0.039    &  0.967  & \multirow{2}{*}{0.560} & \multirow{2}{*}{$> .050$} \\ \cline{2-4}
                                  & high-level ToM     & 0.037     & 1.008  &                        &                        \\ \hline                                   
\multirow{2}{*}{Safety}       & low-level ToM     & -0.118   &  1.108  & \multirow{2}{*}{3.224} & \multirow{2}{*}{$=.016*$} \\ \cline{2-4}
                                  & high-level ToM     & 0.110     & 0.860  &                        &                        \\ \hline                                   

\end{tabular}}
\end{table}

\section{CONCLUSIONS AND FUTURE WORKS}\label{se:conclusion}

In this paper, we investigate whether the ToM abilities of a robot would affect human trust towards the robot. To this end, a Price Game was introduced between human participants and a humanoid Pepper robot. Participants were asked to judge the price of a list of objects, to which the robot could agree or not. In case of disagreement, participants had the chance to change their decision and accept the robot option. This has been done under the premises that the participants' willingness to change their opinion would be led by their trust in the robot knowledge. Therefore, accepting the robot suggestion and changing the price judgement, could be an implicit measure of trust in HRI. Exposing participants to a more or less ToM-skilled robot could, in fact, influence their expectations on the robot subsequent price judgement skills. In other words, if a robot is introduced as not able to emphasise and understand others' point of view, participants might be led to trust less the robot judgement competences.
Our experimental choice was also dictated by previous evidences of trust-related behaviours by using the Price Game \cite{zanatto2019generalisation, zanatto2016priming}.
In order to assess the role of ToM in trust-related behaviours toward robots, we exposed our participants to a video before starting the Price Game. The video could show the same robot used in the Price Game possessing high or low ToM abilities.
Experimental hypotheses gravitated around the idea that being introduced to a robot that shows low ToM ability would degrade participants' trust over its price judgements. 
Results confirmed the experimental hypothesis. Participants changed their judgement more often after being exposed to a high-level ToM video.

It could be then argued that human trust in robots can be affected by the robot ToM abilities. 
The participants' willingness to change their mind on the basis of the robot suggestion, in fact, has been previously used as an implicit measure of the robot acceptance and trust by Zanatto et al~\cite{zanatto2016priming, zanatto2019generalisation}. Although the previous studies focused on the potential  of the robot physical features in activating a trust-related stereotype, our study used the same approach to investigate the role of socio-cognitive-related stereotypes in improving or reducing trust in HRI.
Here, our study gives further consistency to previous findings and also sheds light on the potential effects that applications of ToM in HRI could have. The results here reported showed that a robot possessing a higher level of ToM could be a source of trust, which in turn could improve acceptance and cooperation in HRI. Therefore, ToM could be used in the future to improve the quality of HRI in several contexts, from education to healthcare etc. For instance, if a robot is able to adopt the  elderly point of view, it could be easier for the elderly to comply with requests and suggestions from the robot. Similarly, if the robot can understand the desires or intentions of the students, students are more willing to engage in the teaching.

These results are also partially supported by the questionnaires. A robot with a higher level of ToM was rated more credible, anthropomorphic and safe.  Although not all the questionnaires showed a significant effect of the ToM level, this difference between the implicit measure of
trust and the explicit post hoc measures is not new to the field~\cite{hofmann2005meta}. Moreover, this disparity is partially in line with~\cite{zanatto2019generalisation}. Furthermore, this also remarks how questionnaires might not always capture attitudes and perceptions that the human has not awareness.
Nevertheless, a robot judged as more credible, anthropomorphic and safe, can be in turn entitled of higher trust. These are all features that could be related to the attribute and performance-based features that Hancock lists as affecting trust in HRI.

 
An open question remains evaluating whether this effect could be expanded to other robotic exemplars. As \cite{zanatto2019generalisation,zanatto2016priming} showed that expectations and stereotype activation toward a robot could be transposed to other similars, activations of ToM stereotypes could also be transposed to other types of robots. 

Furthermore, in this study gaze engagement has been introduced to all participants, whereas in previous experiments participants would face also a robot with a fixed gaze over the table. As this manipulation has been found to have an effect on participants' trust rate, it would be advisable, in the future, to investigate the role of ToM on robots showing also different degrees of social gaze.  

Overall, this study confirmed that manipulation of human expectations toward a robot can affect their subsequent interaction with that robot.
Although further investigations are necessary, our results give important contribution to the field, by opening a new and challenging path for the investigations of trust in HRI. This could also deliver new direction in terms of design and bring beneficial effects to several HRI environments in which a more empathetic and socially competent robot would further improve its acceptance and mutual cooperation with humans.






\section*{ACKNOWLEDGEMENT}
This material is based upon work supported by the Air Force Office of Scientific Research, USAF under Award No. FA9550-19-1-7002. The work of Debora Zanatto was funded and delivered in partnership between the Thales Group and the University of Bristol, and with the support of the UK Engineering and Physical Sciences Research Council Grant Award EP/R004757/1 entitled ‘Thales-Bristol Partnership in Hybrid Autonomous Systems Engineering (T-B PHASE)’.


\bibliographystyle{IEEEtran}
\bibliography{reference}
\newpage

\section*{APPENDIX}
\subsection{ToM video scripts.}

The video where the robot fails the false-belief test:

\begin{itemize}
    \item Sam:Hello, Pepper. I am Sam.
    \item Anne: And I am Anne.
    \item The robot: Nice to meet you!
    \item Anne: It is nice to meet you too! OK, I am just gonna put this cube under my green cup.
    \item The robot: OK.
    \item Anne: OK, I have just got to pop out. I will be back in a minute.
    \item The robot: See you!
    \item Anne: See you!
    \item Anne left the room.
    \item Sam: Now that she is gone. I am going get the cube from her cup and I am going to move it under my cup.
    \item The robot: OK.
    \item Sam: Where do you think she will look for the cube when she is back?
    \item The robot: Wait. Let me think. Well. The cube is under the blue cup. So she is going to search there.
    \item Anne comes back...
    \item Anne: Sorry about that.
    \item Anne opens her green cup.
    \item Anne: Oh where is my cube gone?
\end{itemize}

The video where the robot passes the false-belief test:

\begin{itemize}
    \item Sam:Hello, Pepper. I am Sam.
    \item Anne: And I am Anne.
    \item The robot: Nice to meet you!
    \item Anne: It is nice to meet you too! OK, I am just gonna put this cube under my green cup.
    \item The robot: OK.
    \item Anne: OK, I have just got to pop out. I will be back in a minute.
    \item The robot: See you!
    \item Anne: See you!
    \item Anne left the room.
    \item Sam: Now that she is gone. I am going get the cube from her cup and I am going to move it under my cup.
    \item The robot: OK.
    \item Sam: Where do you think she will look for the cube when she is back?
    \item The robot: Wait. Let me think. Well. She did not see that you moved the cup. So she is going to search under the green cup.
    \item Anne comes back...
    \item Anne: Sorry about that.
    \item Anne opens her green cup.
    \item Anne: Oh where is my cube gone?
\end{itemize}


\end{document}